\documentclass[sigconf, nonacm, colorlinks=true,linkcolor=blue]{acmart}

\usepackage{balance}

\pdfoutput=1
\usepackage{hyperref}
\usepackage[noabbrev,capitalise,nameinlink]{cleveref}
\hypersetup{colorlinks=true,linkcolor=blue}
\usepackage{microtype}
\usepackage{graphicx}
\usepackage{booktabs} 
\usepackage{bm}
\usepackage{bbm}
\usepackage{amsmath}  
\usepackage{makecell}
\usepackage{amsthm}
\usepackage{xcolor}
\usepackage{textcomp}
\usepackage{multirow}
\usepackage{mathtools}
\usepackage{geometry}
\usepackage{listings}
\usepackage{boldline}
\usepackage{tabularx}
\usepackage[export]{adjustbox}
\usepackage{mathtools}
\usepackage{graphicx}
\usepackage{xspace}

\newcommand{\gpt}{\textsc{GPT-3}\xspace}

\definecolor{mygray}{gray}{0.6}
\definecolor{keywordcolor}{rgb}{0.2,0.2,0.6}
\definecolor{keywordcolor2}{rgb}{0.15,0.46,0.1}
\definecolor{typecolor}{rgb}{0.17,0.56,0.68}
\definecolor{commentcolor}{gray}{0.3}
\definecolor{ratecolor}{rgb}{0.5,0.1,0.1}
\definecolor{stringcolor}{gray}{0.3}

\usepackage{bm}
\usepackage{enumitem}
\usepackage{caption}
\usepackage{subcaption}

\newcommand*{\ShowNotes}{}

\ifdefined\ShowNotes
  \newcommand{\colornote}[3]{{\color{#1}\bf{#2 #3}\normalfont}}
\else
  \newcommand{\colornote}[3]{}
\fi

\definecolor{darkred}{rgb}{0.7,0.1,0.1}
\definecolor{darkgreen}{rgb}{0.1,0.5,0.1}
\definecolor{cyan}{rgb}{0.7,0.0,0.7}
\definecolor{dblue}{rgb}{0.2,0.2,0.8}
\definecolor{maroon}{rgb}{0.76,.13,.28}
\definecolor{burntorange}{rgb}{0.81,.33,0}
\definecolor{royalpurple}{rgb}{0.47,.31,0.66}

\newcommand {\CITE}[1]{\colornote{magenta}{CITE}{}}
\newcommand {\tasks}[1]{data cleaning and integration }
\ifdefined\ShowNotes
  
\else
  
\fi

\newcommand{\para}[1]{\vspace{0.08in}\noindent\textbf{#1 }}
\newif\ifarxiv

\begin{document}


\title{Can Foundation Models Wrangle Your Data?}





\author{Avanika Narayan, Ines Chami$\dagger$, Laurel Orr, Simran Arora, Christopher R\'e}
\affiliation{%
  \institution{Stanford University and $\dagger$Numbers Station}
  \streetaddress{}
  \city{}
  \state{}
  \postcode{}
  \country{}
}
\email{{avanika, lorr1, chrismre, simarora}@cs.stanford.edu, ines.chami@numbersstation.ai}

\date{}

\nocite{*}

\begin{abstract}
Foundation Models (FMs) are models trained on large corpora of data that, at very large scale, can generalize to new tasks without any task-specific finetuning. 
As these models continue to grow in size, innovations continue to push the boundaries of what these models can do on language and image tasks.
This paper aims to understand an underexplored area of FMs: classical data tasks like cleaning and integration.
As a proof-of-concept, we cast five data cleaning and integration tasks as prompting tasks and evaluate the performance of FMs on these tasks.
We find that large FMs generalize and achieve SoTA performance on data cleaning and integration tasks, even though they are not trained for these data tasks. 
We identify specific research challenges and opportunities that these models present, including challenges with private and domain specific data, and opportunities to make data management systems more accessible to non-experts. We make our code and experiments publicly available at: \url{https://github.com/HazyResearch/fm_data_tasks}.

\end{abstract}

\maketitle



\section{Introduction}

\label{sec:intro}

\begin{figure}[t]
    \centering
    \includegraphics[width=0.40\textwidth]{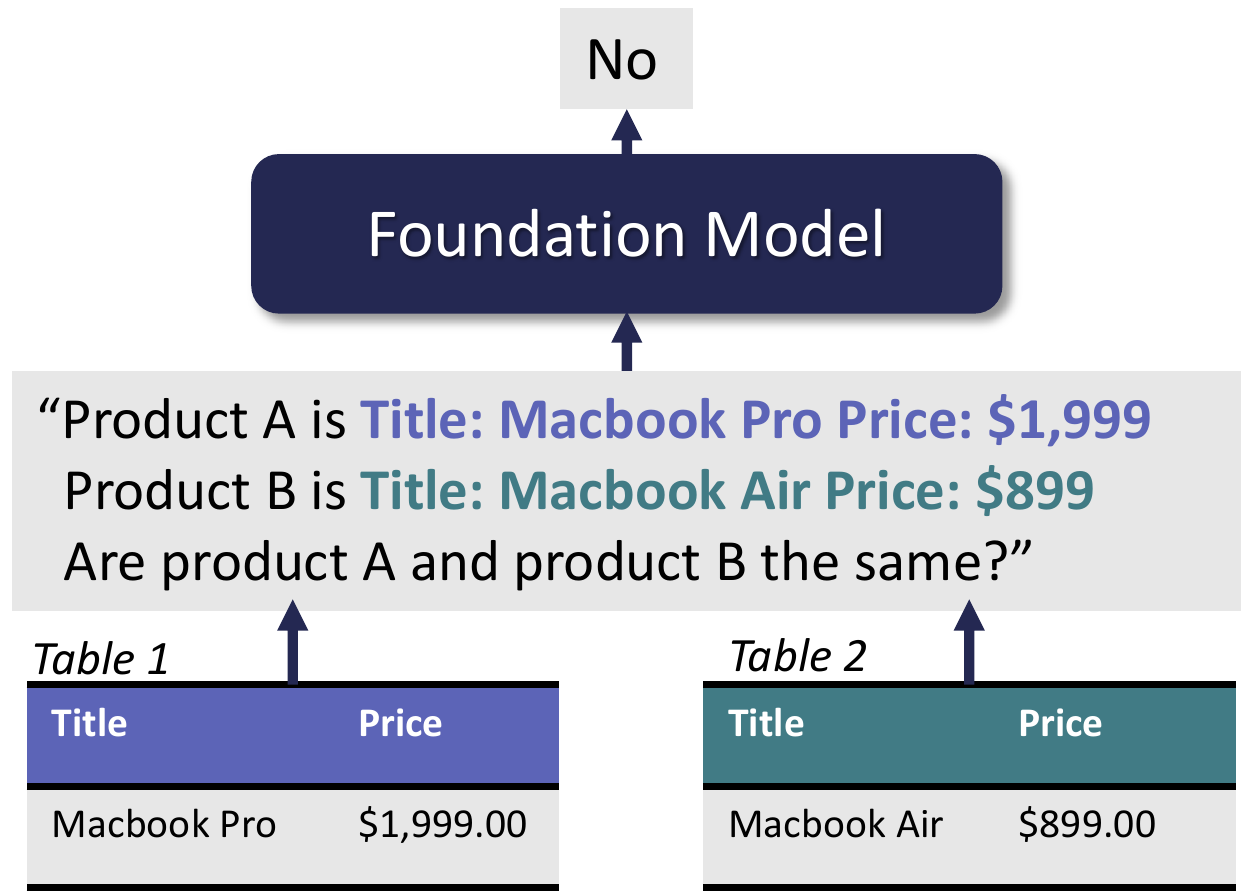}
    \caption{
    A large FM can address an entity matching task using prompting. Rows are serialized into text and passed to the FM with the question ``Are products A and B the same?''.
    The FM then generates a string ``Yes'' or ``No'' as the answer. 
    }
    \label{fig:banner}
\end{figure}
Foundation Models (FMs)~\citep{bommasani2021opportunities} are models trained on broad data that can be adapted to a wide range of downstream tasks.
These models have achieved substantial gains across many semantically challenging tasks such as question answering~\citep{brown2020language}, knowledge base construction~\cite{petroni2019language}, and information retrieval~\cite{guu2020realm}. 
As they have scaled to hundreds of billions of parameters~(e.g. \gpt~\citep{brown2020language}, PaLM~\citep{chowdhery2022palm}), large FMs have demonstrated surprising emergent behaviors and good zero-shot generalization to new tasks (i.e. no task-specific finetuning) on domains vastly different from the data they were pretrained on~\citep{chowdhery2022palm}. 
These large FMs are often autoregressive language models (e.g. \gpt and PaLM) that are trained to predict the next word in large text corpora and can be adapted to new tasks given a simple natural language description of the task~(see \cref{fig:banner}).
These breakthrough capabilities have led to a race for building bigger and better models, and innovations continue to push the boundaries of what large FMs can do on a variety of hard \textit{language tasks}.

A natural question that arises is whether these advances can benefit hard classical \textit{data tasks} (e.g. data cleaning and integration).
While it is clear that FMs benefit text-intensive tasks, it is not clear whether these models can be applied to data tasks over structured data. 
The symbols commonly found in structured data (e.g. dates, numbers, alphanumeric codes) are less frequent in natural language text so it is unclear that FMs possess the ability to reason over them. Moreover, since FMs are trained to predict the next word, it is non-obvious that they can work out-of-the-box on complex data tasks. 
This paper explores the aforementioned question and introduces a new research vision for leveraging FMs for data management, focusing on data cleaning and integration tasks---two keys steps in data-driven enterprise pipelines. 



Recently, a large body of research has applied machine learning (ML)~\cite{konda2016magellan} and deep learning (DL)~\cite{li2020deep, mudgal2018deep} methods---namely pretrained language models (PLMs) like BERT~\citep{devlin2018bert}---to semantically-complex data tasks. 
However, these approaches still require a significant amount of engineering effort as they rely on: 

\begin{itemize}[leftmargin=*]
    \item \textbf{Task-specific architectures}: 
    Data cleaning and integration encapsulate \textit{many different tasks} such as entity matching~\cite{papadakis2020blocking}, schema matching~\cite{sutanta2016survey}, and error detection~\citep{heidari2019holodetect}. 
    Existing approaches, whether they are rule-, ML- or DL-based, vary greatly from one task to the other, often with complex, task-specific architectures. 
    For instance, adapting BERT to data tasks requires architectural changes and finetuning the entire model for each task. This leads to siloed and hard-to-maintain systems.  
    \item \textbf{Hard-coded knowledge}: Data tasks often rely on \textit{domain knowledge} (e.g. understanding the relationship between a city and its zip code for data cleaning constraints) and commonsense reasoning. 
    These are usually hard-coded with human-engineered rules or external knowledge bases \cite{chu2015katara, rekatsinas2017holoclean}. 
    Consequently, systems can be brittle and fail to generalize to a diverse set of domains.
    \item \textbf{Labeled data}: ML- and DL-based solutions require copious amounts of hand-labeled data~\cite{adadi2021survey}. 
    For instance, PLMs that have achieved state-of-the-art (SoTA) results on data tasks (e.g. Ditto~\cite{suchin2020pretraining}) require a significant amount of task-specific labeled data and fine-tuning to achieve good performance. 
    Labeling data for each task is engineering intensive and adds to the difficulty of maintaining data cleaning and integration systems. 
\end{itemize} 

Excitingly, FMs display several useful properties that make them an appealing choice compared to traditional approaches:
\begin{itemize}[leftmargin=*]
    \item \textbf{Task-agnostic architecture}:
    As a result of their natural language interface, FMs can be applied to a wide-range of tasks.  
    For instance,~\cref{fig:banner} shows how an entity matching task---which requires identifying whether two table entries refer to the same entity---can be cast as a prompting task.
    This unifying interface eliminates the need for siloed architectures,
    in contrast to existing learned approaches where architectures need to be carefully crafted for each task (e.g. task-specific classification layer).
    \item \textbf{Encoded knowledge}: Because FMs are trained on large, generic corpora of data, they contain knowledge about an extensive set of {common entities}, and thus do not rely on human-engineered rules to acquire knowledge~\cite{razniewski2021language}. 
    \item \textbf{Limited to no labeled data}: 
    FMs can be applied to a breadth of tasks with little to no labeled data (e.g. few-shot and zero-shot). When a FM needs to be fine-tuned, it typically needs dramatically less labeled data to achieve competitive results~\citep{kaplan2020scaling}.
\end{itemize} 

Our goal is to better understand if large FMs can be applied to data integration and cleaning tasks. 
We study the behavior of \gpt---an early and promising FM.
While \gpt is already a high quality model, we expect the significant investment in FMs from both academia and industry to lead to more performant and scalable FMs over time. 
Like many other communities, the data management community stands to benefit from these trends. 
As such, we aim to understand the advantages and limitations of FMs on data tasks, by focusing on three key questions.

\textit{\textbf{How well do large FMs transfer to data tasks?}}
To answer this, we cast several data tasks as natural language generation tasks~(\cref{sec:approach}) and explore whether a single FM can generalize well to these tasks. 
In~\cref{subsec:fewshot_exp}, we quantify the zero- and few-shot performance of FMs on five enterprise data tasks: entity matching, error detection, schema matching, data transformation, and data imputation. 
We find that the largest \gpt variant (175B parameters) outperforms SoTA ML-and DL-based approaches on these tasks with few examples. This is particularly surprising since prior approaches are fully-finetuned on task-specific labeled data for these tasks, while \gpt-175B is simply pretrained to generate text.

\textit{\textbf{What are the caveats in applying FMs to data tasks?}}
In~\cref{subsec:exp_prompt}, we unpack the few-shot ``prompt tuning'' process---serializing tabular data to text, casting data tasks as text generation tasks and constructing demonstrative task examples---for applying FMs to data tasks.
We quantify the effects of prompt formatting variations on performance and the differences between manually and randomly selecting task examples. 
We find that FMs are brittle to differences in prompt formatting and that performance improves when prompts are manually selected versus randomly selected.


\textit{\textbf{{What opportunities do FMs present for data tasks and what are the relevant research challenges?}}}
Finally, in~\cref{sec:next_steps}, we discuss the potential challenges and related research questions with using FMs in data management pipelines. 
We discuss the forthcoming shift in how ML systems are built, challenges around updating FM knowledge, and opportunities and considerations pertaining to private, temporal and local data. 

We hope that our preliminary exploration will encourage the data management community to explore the effectiveness of FMs for other data tasks and develop techniques to overcome the shortcomings of FMs in this setting.

\begin{figure}[t]
    \centering
    \includegraphics[width=0.47\textwidth]{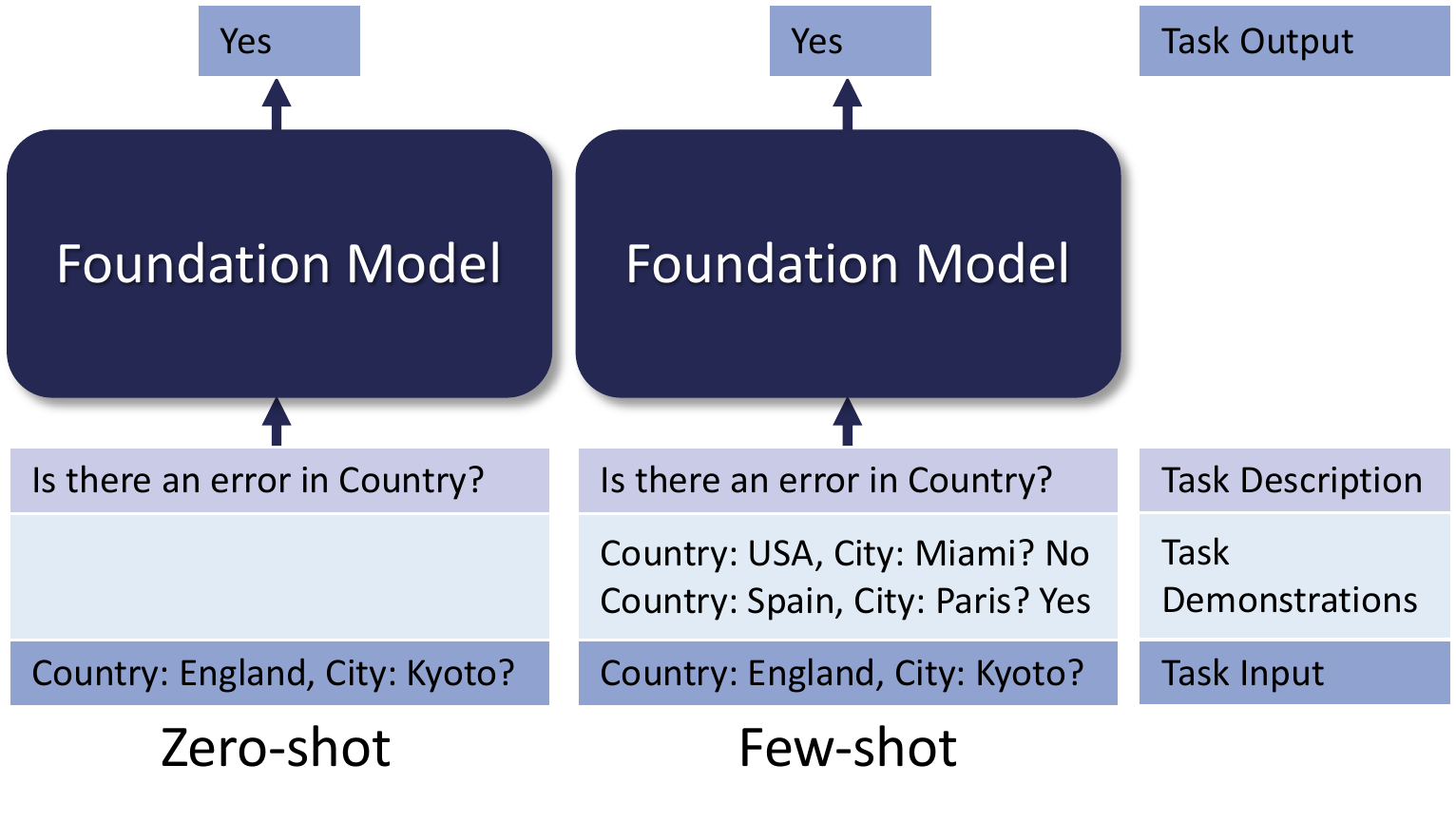}
    \caption{Different ways to use FMs with ``in-context'' learning~\cite{brown2020language} on an error detection task. 
    For zero-shot (left), the prompt is the task description and the example to complete. For few-shot (right), the prompt adds demonstrations of how to complete the task. 
    }\label{fig:fms}
\end{figure}

\section{Background}
We first give some background on the different data tasks considered in this paper and then provide a brief review of FMs.

\subsection{Problem Setup}\label{subsec:tasks}
We focus on entity matching (EM), error detection (ED), and data imputation (DI) and describe the setup for these tasks. We denote $D$, a structured dataset with $n$ entries, such that each entry is represented by a collection of $m$ attribute value pairs: for entry $e_i \in D$ we have $e_i = \{e_{i,j}\}_{1 \leq j \leq m}$ where for attribute $j$, $e_{i,j}=\{\texttt{attr}_j, \texttt{val}_j\}$.

\para{\textbf{Entity Matching}}
The goal of EM is to match entities (real-world objects like people, places and things) across different datasets. Formally, given two structured datasets $(D, D')$ and pairs of entries $e,e'\in D\times D'$, the goal is to predict whether these entries represent the same entity or not. 
This problem is usually solved as a classification problem, and real-world EM systems are often preceded by blocking heuristics which are used to remove obvious non-matches.  

EM has been extensively studied over the past decade (see~\cite{papadakis2020blocking} for a survey) and methods broadly fall into three categories: rule-based, crowd-based~\cite{gokhale2014corleone, wang2012crowder} and ML/DL-based~\cite{konda2016magellan,mudgal2018deep}. 
Recently, methods relying on PLMs~\cite{li2020deep} have become SoTA for this task. 

\para{\textbf{Error Detection}} ED is an important step in data cleaning pipelines.
Given an entry $e$, the goal is to detect attributes $j$ where $\texttt{val}_j$ has an error. 
The task is framed as a classification problem where the goal is to predict if $\texttt{val}_j$ is correct for a given $e$.

ED has been studied extensively in both academic and industrial settings~\citep{abedjan2016detecting}. 
There are a number of successful commercial offerings including Trifacta~\citep{Trifacta} and Tamr~\citep{tamr}. 
Traditionally, ED systems have been heavily reliant on rule-based algorithms which enforce data constraints through functional dependencies or knowledge bases~\citep{chu2015katara,chu2013holistic, dallachiesa2013nadeef}. 
Additionally, there are statistical-based approaches such as pattern enforcement~\citep{kandel2011wrangler, chu2015katara}, outlier detection~\citep{dasu2012statistical}, and record deduplication~\citep{stonebraker2013data} algorithms.
Recent efforts have developed SoTA ML models for ED~\citep{heidari2019holodetect}.

\para{\textbf{Data Imputation}} 
DI is a critical step for repairing dirty data sources.
Given an entry $e$ with missing attribute values $\{\texttt{attr}_j, \texttt{NULL}\}$, the goal of DI is to infer the missing values.
The full range of plausible values for the missing value is not known apriori. 

Prior works in DI falls into three categories---clustering and/or statistical-based~\citep{mayfield2010eracer}, generative model-based ~\citep{ghysels2007midas},  ML/DL-based~\citep{biessmann2019datawig, mei2021capturing}, 
and tabular data pretraining-based approaches~\citep{deng2022turl}---and struggle when needing to impute values not seen in the training set~\citep{mei2021capturing}.

\subsection{\textbf{Background on Foundation Models}}\label{subsec:fm_background}
We now give an overview of language FMs, starting from very early, smaller FMs (i.e. PLMs) and moving to large-scale FMs, the latter of which is the focus of this paper.

\para{Pretrained Language Models}
Pretrained Language Models (PLMs) are neural networks pretrained on large corpora of publicly available text (e.g. web pages). 
The first breed of PLMs---ELMo~\cite{peters-etal-2018-deep}, BERT~\cite{devlin-etal-2019-bert}, RoBERTA~\citep{liu2019roberta}---learned the semantics of natural language by predicting masked works during pretraining. 
These models have on the order of hundreds of millions of parameters. 
Traditionally, PLMs are adapted to downstream tasks through a task-specific prediction layer (e.g. classification layer) and a task-specific finetuning step wherein all model weights are updated. 

\para{\textbf{Large Autoregressive Language Models}}
In 2020, \gpt~\citep{brown2020language} marked a significant shift in the the ML community. 
It represented a new class of large-scale language models: autoregressive language models pretrained to predict the next word in a sequence.
These models have \emph{billions} of parameters and have been used for language generation tasks such as question answering and summarization. Since the release of \gpt, bigger and better performing autoregressive language models have been developed~\citep{chowdhery2022palm}.

\para{\textbf{Emergent Behaviors}}
Interestingly, the {biggest} \gpt variant (175B parameters) has the capacity to solve natural language tasks with only a few examples (few-shot prompting), and, in some cases, just a task description (e.g. ``Translate French to English''). Unlike traditional finetuning, no model parameters are updated to fit the task. Few-shot prompting has proven to be effective on tasks widely different from the FMs pretraining objective (e.g., code generation~\cite{xu2022systematic}, Trivia QA~\cite{lin2021truthfulqa, brown2020language} and arithmetic~\cite{brown2020language}). This in-context learning behavior can be thought of as the FM “locating” an already-learned behavior~\citep{reynolds2021prompt}. Smaller models (<10B parameters) typically require some form of task-specific finetuning.
We study ways to use smaller FMs on data tasks in the full report~\citep{narayan2022can}. 
\section{Foundation Models for Data Tasks}
\label{sec:approach}
Our goal is to understand whether FMs can benefit data cleaning and integration tasks. 
The procedure for applying FMs to data tasks requires adapting structured data inputs to textual inputs~(\cref{subsec:serialization}), casting data tasks as text generation tasks~(\cref{subsec:prompting}) and, for few-shot prompting, constructing demonstrative task examples to help the FM learn new data tasks~(\cref{subsec:few_shot_prompts}). 

\subsection{Tabular Data Serialization}\label{subsec:serialization}
FMs take text as input and generate text as output. 
In order to apply these models to data tasks, we first need to convert structured tabular data inputs to text representations.  

Given a structured dataset, we first convert an entry to text. 
Concretely, for entry \texttt{e}, we follow previous work~\cite{li2020deep} and serialize the attribute value and entry as follows:
\begin{align*}
    &\texttt{serialize}(e) \coloneqq \texttt{attr}_1: \texttt{val}_1 \ldots \texttt{attr}_m: \texttt{val}_m
\end{align*}
If the attribute value is \texttt{NULL}, it is serialized as the empty string.
Based on the task and dataset, serialization only happens over a subset of attributes relevant to the task.
FMs can be sensitive to the prompt format and the specific serialization used~\citep{zhao2021calibrate}. In~\cref{subsec:exp_prompt} we study the sensitivity of FMs to sub-selection choices. 

 

\subsection{Data Tasks as Natural Language Tasks}\label{subsec:prompting}
Next, we need to convert data tasks to text generation tasks.
We construct natural language descriptions of each task (i.e. prompts) that use the serialized representations from ~\cref{subsec:serialization}. 
The prompts are passed to the FM, whose generated output is the answer to the given task. 
For entity matching and error detection, the model generates a ``Yes'' or ``No'' response,\footnote{Interestingly, the model is not constrained to produce a Yes/No answer on the output side, but we find that this happens most of the time. In the rare examples where the model does not predict a Yes/No answer, we use {No} as the default answer.} and for imputation it generates the missing value.
We now enumerate the prompts for each task.\\

\noindent\textit{\underline{Entity matching}}: given two entries $(e, e')$ the template is
\begin{align*}
    &\texttt{Product A is }\texttt{serialize}(e)\texttt{. Product B is }\texttt{serialize}(e').\\
    &\texttt{Are Product A and Product B the same?}
\end{align*}

\noindent\textit{\underline{Data imputation}}: given an entry $e$ and attribute $j$ to infer, we use 
\begin{align*}
    \texttt{attr}_1: \texttt{val}_1 \ldots \texttt{attr}_j\texttt{ ?}
\end{align*}

\noindent\textit{\underline{Error detection}}: given an entry $e$ and attribute $j$ to classify as erroneous, we use
\begin{align*}
    \texttt{Is there an error in } \texttt{attr}_j:\texttt{val}_j\texttt{?}
\end{align*}
These templates highlight the generality of this framework which could be extended beyond the tasks considered in this paper. 



\subsection{Task Demonstrations}\label{subsec:few_shot_prompts}
Task demonstrations can be included in the prompt to help the model learn a new task~(see~\cref{fig:fms} for an error detection example). 
These demonstrations are used to show the model how the task should be completed (e.g. should it generate Yes/No or a missing value) as well as understand the finer-grained semantics of $D$. 
We explore two approaches for selecting task demonstration examples. 

\para{Random}
One approach is to sample random examples from a labeled dataset. 
However, this approach often causes high variance in FM performance~\citep{liu2021makes}. Moreover, the ordering of examples in the prompt can have a non-trivial impact on the downstream performance~\citep{lu2021fantastically,zhao2021calibrate, liu2021makes}. 
Although recent work has tried to systematize the prompt tuning process, it is still an open area of research~\citep{li2021prefix, sun2020conditioned}.

\para{Manual}
Another approach is to manually construct examples that yield good performance on a held-out validation set that is 10\% of the original labeled dataset. 
This approach is more costly (requires more time) in comparison to random sampling but improves performance when examples are carefully constructed.

In our manual prompt tuning experiments, we spend \textit{at most 1 hour} per task analyzing errors on the validation set.
We then manually construct demonstrations that help the model correct these errors.
We liken this step to the canonical hyperparameter tuning process in ML where time and compute are spent finding the best model parameters for the given task. 
However, the engineering effort for prompt tuning is significantly lower.
Concretely, it takes less time (minutes vs. hours and days), is more compute effective (inference vs. full training), and gives the user finer-grained control (natural language guidance vs. blackbox parameter updates).

\begin{table}[t]
    \centering
    \caption{Entity matching results measured by F1 score where \textit{k} is the number of task demonstrations.}
    \resizebox{0.45\textwidth}{!}{\renewcommand{\arraystretch}{0.98}
    \begin{tabular}{ |c||c|c|c|c|}
     \hline
     \multirow{2}{*}{Dataset}        & \multirow{2}{*}{Magellan} &\multirow{2}{*}{Ditto}& GPT3-175B & GPT3-175B \\
     & & & ($k$=0) & ($k$=10) \\
     \hline
     Fodors-Zagats  & 100  & 100 & 87.2 &\textbf{100} \\
     Beer & 78.8 & 94.37 & 78.6 & \textbf{100} \\
     iTunes-Amazon & 91.2 & 97.06 & 65.9 &\textbf{98.2}\\
     Walmart-Amazon & 71.9 & 86.76 & 60.6 & \textbf{87.0}\\
      DBLP-ACM & 98.4 & \textbf{98.99} & 93.5 & 96.6\\
     DBLP-Google    & 92.3 & \textbf{95.60} & 64.6 & 83.8\\
     Amazon-Google  & 49.1 & \textbf{75.58} & 54.3 & 63.5\\
    \hline
    \end{tabular}
} 
\label{tab:global}
\end{table}

\section{Experiments}
\label{sec:exp}
We compare FMs to SoTA methods on a variety of data cleaning and integration tasks. 
Our goal is to understand whether large FMs can transfer to data tasks in zero- and few-shot settings~(\cref{subsec:fewshot_exp}), and the nuances in applying FMs to data tasks~(\cref{subsec:exp_prompt}).

\subsection{Experimental Setup}
We begin by describing our experimental protocol, including models, datasets, metrics and baselines used. 

\para{{Models}} 
For few-shot prompting we use the \gpt-175B parameter model (text-davinci-002) in the OpenAI API endpoint~\cite{openai_2021}. 

\para{{Datasets}} 
For entity matching, we use the standard Magellan benchmark~\citep{konda2016magellan}. For schema matching, we choose a challenging dataset, Synthea, from the OMAP benchmark~\citep{zhang2021smat}. For data transformation, we choose two challenging datasets from the TDE benchmark~\citep{he2018transform}.
For imputation, we choose two challenging datasets from ~\citep{mei2021capturing}: Restaurants and Buy.
Finally, for error detection, we evaluate on the benchmark Hospital and Adult datasets which are used across several data cleaning papers~\citep{heidari2019holodetect, rekatsinas2017holoclean, chu2013holistic}.  

We use the provided train/test/dev splits for all entity matching datasets. 
For cleaning tasks, the splits are not available but we follow the protocol of~\citep{heidari2019holodetect,mei2021capturing} to generate the dataset splits. For the Adult dataset, we evaluate over a randomly sampled set of 1K rows due to budget constraints.

\para{\textbf{Evaluation Metrics}}
For the error detection, and schema/entity matching, we evaluate performance using F1 score. 
For imputation and data transformation, we evaluate using accuracy.


\para{{Baselines}} 
We compare against the SoTA methods for each task. 
For entity matching, we benchmark against Ditto~\citep{li2020deep}, the current SoTA DL-based approach which finetunes BERT~\cite{devlin-etal-2019-bert}. 
For data imputation, we benchmark against IMP~\citep{mei2021capturing}, which finetunes RoBERTa~\citep{liu2019roberta}, and HoloClean~\citep{rekatsinas2017holoclean}, a statistical-based SoTA data repair engine. For schema matching, we compare against the SoTA model, SMAT~\citep{zhang2021smat}, which finetunes an attention-based BiLSTM. For data transformations, we compare against TDE~\citep{he2018transform}, a SoTA search-based solution.
Finally, for error detection, we compare against HoloClean and HoloDetect~\cite{heidari2019holodetect}, a data-augmentation based ML approach. 

\begin{table}[t]
    \centering
    \caption{
Data cleaning results, measured in accuracy for data imputation and F1 score for error detection where \textit{k} is the number of task demonstrations.} 
    \resizebox{0.45\textwidth}{!}{\renewcommand{\arraystretch}{0.98}
    \begin{tabular}{|c||c|c|c|c|c|c|c|}
     \hline
     Task & \multicolumn{2}{c|}{Imputation} & \multicolumn{2}{c|}{Error Detection}  \\
     \hline
     Dataset & Restaurant & Buy & Hospital & Adult \\
     \hline
     HoloClean & 33.1 & 16.2 & 51.4 & 54.5 \\
     IMP & 77.2 & 96.5  & - & - \\
     HoloDetect & - & - & 94.4 & 99.1 \\
     GPT3-175B ($k$=0) & 70.9 & 84.6 & 6.9 & 0.0\\
     GPT3-6.7B ($k$=10) & 80.2 & 86.2 & 2.1 & 99.1 \\
     GPT3-175B ($k$=10) & \textbf{88.4} & \textbf{98.5} & \textbf{97.8} & \textbf{99.1} \\
     \hline
    \end{tabular}
    }
\label{tab:global_imp}
\end{table}


\vspace{10mm}
\begin{table}[t]
    \centering
    \caption{
Data integration results, measured in accuracy for data transformations and F1 score for schema matching. Previous SoTA method is TDE for data transformation and SMAT for schema matching.}
    \resizebox{0.45\textwidth}{!}{\renewcommand{\arraystretch}{0.98}
    \begin{tabular}{|c||c|c|c|}
     \hline
     Task &  \multicolumn{2}{c|}{Data Transformation} & \multicolumn{1}{c|}{Schema Matching} \\
     \hline
     Dataset & StackOverflow & Bing-QueryLogs & Synthea\\
     \hline
     Previous SoTA & 63.0 & 32.0 & 38.5 \\
     GPT3-175B ($k$=0) & 32.7 & 24.0 & 0.5\\
     GPT3-175B ($k$=3) & \textbf{65.3} & \textbf{54.0} & \textbf{45.2}\\
     \hline
    \end{tabular}    }
\label{tab:global_di}
\end{table}
\vspace{-10mm}
\subsection{Zero/Few-shot Performance of Large FMs}\label{subsec:fewshot_exp}
In this section we explore the zero and few-shot performance of \gpt-175B. 
Our goal is to understand whether large FMs transfer to data tasks in zero- and few-shot settings. 

\subsubsection{Results}
We first review the zero- and few-shot results.

\para{Few-shot Performance} 
Our results show that in the few-shot setting, with manually curated task demonstrations, \textit{\gpt-175B achieves SoTA performance on 4 entity matching, 2 imputation, 2 data transformation, 2 error detection and 1 schema matching benchmark dataset(s)}~(see \cref{tab:global}, \cref{tab:global_imp}, \cref{tab:global_di}). 
The FM outperforms fully-finetuned, SoTA PLM-based approaches for entity matching~\citep{li2020deep}, and data imputation~\citep{mei2021capturing}. 
For error detection, the few-shot approach outperforms the ML-based SoTA method, HoloDetect, which uses data augmentation and weak supervision to perform well. 

\para{Zero-shot Performance}
In the zero-shot setting, we observe that the FM outperforms statistical-based approaches and standard data repair engines~\citep{rekatsinas2017holoclean} for imputation.
On entity matching, the zero-shot performance is significantly lower than the few shot performance.
This performance gap suggests that demonstrations are very important for the task, and we study the impact of task demonstrations in more detail in~\cref{subsec:exp_prompt}.

\subsubsection{Discussion}
These results show that large FMs can transfer to data tasks. 
These results are particularly exciting given that FMs are trained to model English language and have no prior exposure to the semantics of data tasks nor the syntax of tabular data.
Furthermore, the zero-shot performance on imputation suggests that large FMs not only have an understanding of how to complete the tasks, but also have encoded \textit{knowledge} that is needed to correct and complete records (e.g. functional dependencies between address and zip code). We analyze encoded large FM knowledge in the full report~\citep{narayan2022can}. 

On the entity matching datasets that the FM does not achieve SoTA on, we find that the FM struggles on data domains that contain jargon not commonly found in text. In such cases, the model lacks a strong semantic understanding of the input and has difficulty reasoning over the data. 
For example, in the Amazon-Google dataset, the model has difficulty matching samples due to the high volume of product-specific identifiers in the descriptions. 
Here, for instance, the model fails to accurately match the two entries: \textit{``name: pcanywhere 11.0 host only cd-rom xp 98 nt w2k me. manufacturer: symantec. price: NULL''} and \textit{``name: symantec pcanywhere 11.0 windows. manufacturer: NULL. price: 19.99.''}. 
We discuss ways to adapt FMs to domain-specific data in more detail in~\cref{sec:next_steps}.

\begin{table}[t]
    \centering
    \caption{Entity matching ablation results (F1 score) for different prompt formats ($k$=10). For all datasets, we evaluate on up to 200 samples for cost purposes.} 
    \resizebox{0.47\textwidth}{!}{\renewcommand{\arraystretch}{0.98}
    \begin{tabular}{ |l||c|c|c|}
     \hline
     \multirow{2}{*}{\textbf{Prompt Format}} & \multirow{2}{*}{\textbf{Beer}} & {\textbf{iTunes-}} & {\textbf{Walmart-}} \\
     & & \textbf{Amazon} & \textbf{Amazon} \\
     \hline
     Prompt 1 (w. Attr. \& Example Select.) & \textbf{100} $\pm$ 0.00  & \textbf{98.2} $\pm$ 0.00  & {88.9} $\pm$ 0.00   \\
     Prompt 1 (w/o Example Select.) & 91.1 $\pm$ 0.05 & 86.6 $\pm$ 0.02 & 65.2 $\pm$ 0.04\\
     Prompt 1 (w/o Attr. Select.) &76.9 $\pm$ 0.00 & 94.1 $\pm$ 0.00 & 75.0 $\pm$ 0.00 \\
     Prompt 1 (w. Attr. \& w/o Attr. names)  & 80.0 $\pm$ 0.00  & 94.5$\pm$ 0.00 & 84.2 $\pm$ 0.00 \\
     Prompt 2 (w. Attr. \& Example Select.)  & 96.3 $\pm$ 0.00  & 84.7 $\pm$ 0.00 & \textbf{100} $\pm$ 0.00 \\
     \hline
        \multicolumn{4}{|l|}{Prompt 1: ``\texttt{Are Product A and Product B the same?}''} \\
        \multicolumn{4}{|l|}{Prompt 2: ``\texttt{Are Product A and Product B equivalent?}''}\\
    \hline
    \end{tabular}}
\label{tab:prompt_ablation}
\end{table}
\subsection{Prompt Tuning Ablations}
\label{subsec:exp_prompt}
In this section, we analyze the performance impact of the three different choices made during prompt tuning: attribute selection, prompt formatting, and task demonstration curation. 

\subsubsection{Results}
We run our ablations on three entity matching datasets (see \cref{tab:prompt_ablation}). For all datasets, we evaluate on up to 200 samples for cost purposes. We discuss our findings next.

\para{\textbf{Attribute Selection}}
First, we find through experimentation that sub-selecting attributes during row serialization can have a non-trivial impact on entity matching performance. 
In particular, we observe that sub-selecting attributes that are essential in determining whether two entities match (e.g. name) and removing noisy attributes improves model accuracy. 
To better illustrate this point, we evaluate FM performance on three datasets when not sub-selecting attributes. We find that, on average, attribute selection results in a 13.7 F1 point performance improvement~(\cref{tab:prompt_ablation}). 

\para{\textbf{Prompt Formatting}}
Second, we observe that FMs can be brittle to subtle variations in prompt templates. We investigate this brittleness by replacing the span ``\texttt{Are Product A and Product B the same?}'' (Prompt 1) with ``\texttt{Are Product A and Product B equivalent?}'' (Prompt 2). This minor modification results in an average 9.4 F1 point performance gap in the datasets in~\cref{tab:prompt_ablation}. 

\para{\textbf{Task Demonstrations}}
Finally, we find that the choice of task demonstrations has a significant impact on downstream performance. 
We conduct an ablation where we replace manually curated task demonstrations with randomly selected demonstrations (see Prompt 1 (w/o Example Select.) in ~\cref{tab:prompt_ablation}). We run this experiment over three different random seeds and report the results in~\cref{tab:prompt_ablation}.
Across all cases, manually curated examples outperform randomly selected examples by an average of 14.7 F1 points. 

\vspace{-2pt}
\subsubsection{Discussion}
The aforementioned results demonstrate that successful prompt tuning requires (1) selecting an informative set of attributes required for the task, (2) crafting a well-formatted prompt that the FM understands, and (3) constructing a set of instructive task demonstrations that condition the model to the data at hand. For (1), we find that attribute sub-selection boosts performance by removing noisy attributes that hurt performance. For (2), our ablations show that prompt formatting (e.g. word choice, punctuation) can have significant impact on model performance. For (3), our results indicate that examples need to be carefully crafted for FMs to learn new tasks. 
We conjecture that on more reasoning-intensive tasks (e.g. matching), prompts are important as they help teach the model how to reason about entities and how to complete the task. Moreover, we emphasize that all three steps require some form of iteration to develop the most effective prompt (e.g. passing various inputs to the model and inspecting its outputs).

These findings are aligned with existing literature on prompt tuning that observe non-trivial variance in prompt-based learning settings~\citep{zhao2021calibrate}.  
This performance variance suggests that iterative prompt programming is an essential human-in-the-loop process for FM usage~\citep{liu2021pre}. However, some works suggest that smarter, automatic example selection methods can help close the gap between random example selection and human-in-the-loop prompt selection~\citep{liu2021makes}. These results highlight the paradigm shift induced by building systems centered around FMs: instead of spending time tuning models, we now need to invest time finding the right examples and engineering useful prompts for each task.
We discuss these paradigm shifts in more detail in~\cref{sec:next_steps}.

\section{Research Agenda}
\label{sec:next_steps}
Because of their natural language interface and vast internal knowledge, FMs provide an interface for unifying a wide-range of siloed, hand-engineered data integration pipelines. Consequently, \textit{we envision that the data orchestration workbenches of the future will be centered around FMs}. We discuss the opportunities, practical considerations, and technical challenges associated of this vision next.





\subsection{Opportunities for FMs}
\label{sec:opportunities}


\para{Natural Language Interactions} FMs usher in a new era of human-machine collaborative workflows wherein users spend less time labeling data and finetuning models and more time writing natural language prompts that are representative of the task at hand. As the necessity of writing code decreases, we envision systems that are more accessible to non-machine learning experts (e.g., business users). In future work, we seek to better understand the human-in-the-loop prompt engineering pipeline, especially in the context of data management practices.

\para{Model Prototyping} The data integration and management pipeline can be categorized into three distinct stages: discovery and design, development, and deployment~\citep{data_integration_phases}. We propose that FMs will be most useful in the discovery and design phase when training data is less available. In this setting, FMs enable rapid prototyping of data models via prompting. In some cases, the FM's out-of-the-box performance will be sufficiently high, obviating the need to train a task-specific model. In others, we can use the FM to label and generate data with human-in-the-loop feedback. When a sufficient amount of data has been collected, transitioning to the fully-supervised model development regime is the optimal choice.



\para{Passive Learning From User Exhaust} In the enterprise setting, organizations accumulate a staggering amount of data exhaust---the informational byproduct that streams from devices, products, and workforce management practices~\citep{dataexhaust}. Because FMs are pretrained in an unsupervised fashion with a simple token prediction objective (see Section~\ref{subsec:fm_background}), they can learn over \textit{any} raw and unlabeled sources of data~\citep{DBLP:journals/corr/abs-2005-00341, yan2021videogpt, reed2022generalist}. As such, FMs can effectively ingest the exhaust from the 
entire data stack (from system logs to structured data). Learning from data analyst exhaust (e.g., clicks over GUI) is also an opportunity to improve FM performance for these data tasks.
\subsection{Practical Considerations of FMs}
\label{sec:practical_considerations}

\para{\textbf{Integration in Data Management Workflows}} 
FMs take text as input and generate text as output. As a result, they are limited in their ability to directly take actions over the graphical user interfaces (GUIs) of data management software (DMS) (e.g., Snowflake). Given that a majority of data analyst time is spent working in these environments, we need ways of translating natural language specifications of data tasks (e.g., “remove all nan cells”), to actionable operations on these GUIs. Excitingly, new work demonstrates how to augment FM capabilities such that they can effectively take actions on web and mobile interfaces~\citep{wang2022enabling, adept_act1}. These works suggest the possibility of directly integrating FMs with DMS.

\para{\textbf{Interaction with Existing Systems}}
FMs can interact with existing DMS by either replacing them or utilizing their outputs to conduct downstream tasks. In terms of system replacement, for tasks where the required rules and logic are not encoded in the FMs knowledge, it is an open question~\citep{desmond2022no} as to how to systematically translate the rules and logic (e.g., domain-specific dependencies) to natural language inputs to the FMs. For system integration, we need ways of systematically incorporating the outputs of existing systems (e.g., dataset pattern discovery) in natural language prompts. Prior work has demonstrated how to ``ground'' FM-based text-to-SQL prompts with database schemas and metadata~\citep{dengstructure}. Similar ideas could be used when incorporating system outputs with FMs.


\para{\textbf{Debuggability}}
FMs are non-deterministic and can make unexpected errors. In order to use these models in data management pipelines, we need mechanisms for increasing transparency and debuggability of pipelines. One possible approach is to collect and monitor model confidence scores. Prior work demonstrates that a FM can “learn to express uncertainty about its own answers in natural language”~\citep{lin2022teaching}. Another approach is to decompose tasks into chains of “primitive operations”, enabling for more transparency and visibility of specific failure points~\citep{wu2022ai}.

\subsection{Technical Challenges of FMs}

\label{sec:challenges}
\para{\textbf{Domain Specificity}}
A key challenge in applying FMs to data management tasks is operating over highly specialized domains (e.g., medical, financial, and insurance data). The existing literature suggests that domain-specialization is best achieved by continuously pretraining on relevant data (e.g., earning reports, medical reports) ~\cite{yang2020finbert, gururangan2020don}. However, training models with billions of parameters can be costly. This has motivated several works which attempt to more quickly and cheaply adapt models to new domains by finetuning only a few layers of the model or simply training a small neural network (e.g., adaptor) on top of a frozen FM~\citep{frozen2022}.

\para{\textbf{Privacy}}
Organizations are often unable to pass sensitive data to third party APIs for privacy reasons~\citep{arora2022can, arora2022reasoning}. Unfortunately, this constraint is in conflict with the contemporary state of FMs where the best performing models (e.g., GPT-3), can only be accessed via API requests. This motivates the need for better open-source models which are competitive with closed-source models. Excitingly, recent work~\citep{arora2022ask} has demonstrated that techniques such as prompt ensembling and prompt reframing can enable open-source models like GPT-J-6B~\citep{gpt-j} to out-perform GPT3-175B on popular natural language understanding benchmarks. Extending this work to data management tasks is an interesting research direction.

\para{\textbf{Prompt Engineering and Automation}}
Prompting is the primary mechanism for adapting FMs to new tasks. However, prompting requires some manual effort to design performant prompts. In the DI setting, prompts can be sensitive to things such as data schemas and minor formatting variations. A few automated approaches have been developed which reduce the manual effort needed to construct prompts: soft prompt tuning (e.g., optimizing a sequence of task-specific vectors which are appended to the text prompt)~\citep{lester-etal-2021-power, li-liang-2021-prefix} and learning to retrieve better in-context examples~\citep{rubin2021learning}. Adapting these works to the challenges of prompting for data tasks is an open area of research.

\section{Conclusion}\label{sec:conc}
In this work we investigate the applicability of FMs to classical data tasks. We find that large FMs can achieve SoTA performance on many data tasks with 0-to-few natural language task demonstrations. \textit{The ability of these models to transfer to data tasks with no task-specific finetuning is particularly interesting given that these models are simply trained to predict next words}. Our work builds upon years of important work on integration and cleaning tasks in the data management community. We hope that our results gesture towards the possibilities of using language guided models for human-in-the-loop data integration practices across a broader range of data management tasks.
\section*{Acknowledgments}

We are thankful to Ihab Ilyas, Theo Rekatsinas, Mike Cafarella, Ce Zhang, Sen Wu, Christopher Aberger, Neel Guha, Beidi Chen and Xiao Ling for their helpful discussions and feedback. We gratefully acknowledge the support of DARPA under Nos. FA86501827865 (SDH) and FA86501827882 (ASED); NIH under No. U54EB020405 (Mobilize), NSF under Nos. CCF1763315 (Beyond Sparsity), CCF1563078 (Volume to Velocity), and 1937301 (RTML); ONR under No. N000141712266 (Unifying Weak Supervision); the Moore Foundation, NXP, Xilinx, LETI-CEA, Intel, IBM, Microsoft, NEC, Toshiba, TSMC, ARM, Hitachi, BASF, Accenture, Ericsson, Qualcomm, Analog Devices, the Okawa Foundation, American Family Insurance, Google Cloud, Swiss Re,
Brown Institute for Media Innovation,
Department of Defense (DoD) through the National Defense Science and
Engineering Graduate Fellowship (NDSEG) Program, 
Fannie and John Hertz Foundation,
National Science Foundation Graduate Research Fellowship Program,
Texas Instruments Stanford Graduate Fellowship in Science and Engineering,
and members of the Stanford DAWN project: Teradata, Facebook, Google, Ant Financial, NEC, VMWare, and Infosys. The U.S. Government is authorized to reproduce and distribute reprints for Governmental purposes notwithstanding any copyright notation thereon. Any opinions, findings, and conclusions or recommendations expressed in this material are those of the authors and do not necessarily reflect the views, policies, or endorsements, either expressed or implied, of DARPA, NIH, ONR, or the U.S. Government.

\ifarxiv
\section*{Acknowledgments}

We are thankful to Ihab Ilyas, Theo Rekatsinas, Mike Cafarella, Ce Zhang, Sen Wu, Christopher Aberger, Neel Guha, Beidi Chen and Xiao Ling for their helpful discussions and feedback. We gratefully acknowledge the support of DARPA under Nos. FA86501827865 (SDH) and FA86501827882 (ASED); NIH under No. U54EB020405 (Mobilize), NSF under Nos. CCF1763315 (Beyond Sparsity), CCF1563078 (Volume to Velocity), and 1937301 (RTML); ONR under No. N000141712266 (Unifying Weak Supervision); the Moore Foundation, NXP, Xilinx, LETI-CEA, Intel, IBM, Microsoft, NEC, Toshiba, TSMC, ARM, Hitachi, BASF, Accenture, Ericsson, Qualcomm, Analog Devices, the Okawa Foundation, American Family Insurance, Google Cloud, Swiss Re,
Brown Institute for Media Innovation,
Department of Defense (DoD) through the National Defense Science and
Engineering Graduate Fellowship (NDSEG) Program, 
Fannie and John Hertz Foundation,
National Science Foundation Graduate Research Fellowship Program,
Texas Instruments Stanford Graduate Fellowship in Science and Engineering,
and members of the Stanford DAWN project: Teradata, Facebook, Google, Ant Financial, NEC, VMWare, and Infosys. The U.S. Government is authorized to reproduce and distribute reprints for Governmental purposes notwithstanding any copyright notation thereon. Any opinions, findings, and conclusions or recommendations expressed in this material are those of the authors and do not necessarily reflect the views, policies, or endorsements, either expressed or implied, of DARPA, NIH, ONR, or the U.S. Government.

\fi

\bibliographystyle{ACM-Reference-Format}
\bibliography{main}

\pagebreak
\textcolor{white}{THIS IS DUMMY TEXT}
\vspace{4cm}
\pagebreak
\appendix

\begin{figure}[t]
    \centering
    \includegraphics[width=0.35\textwidth]{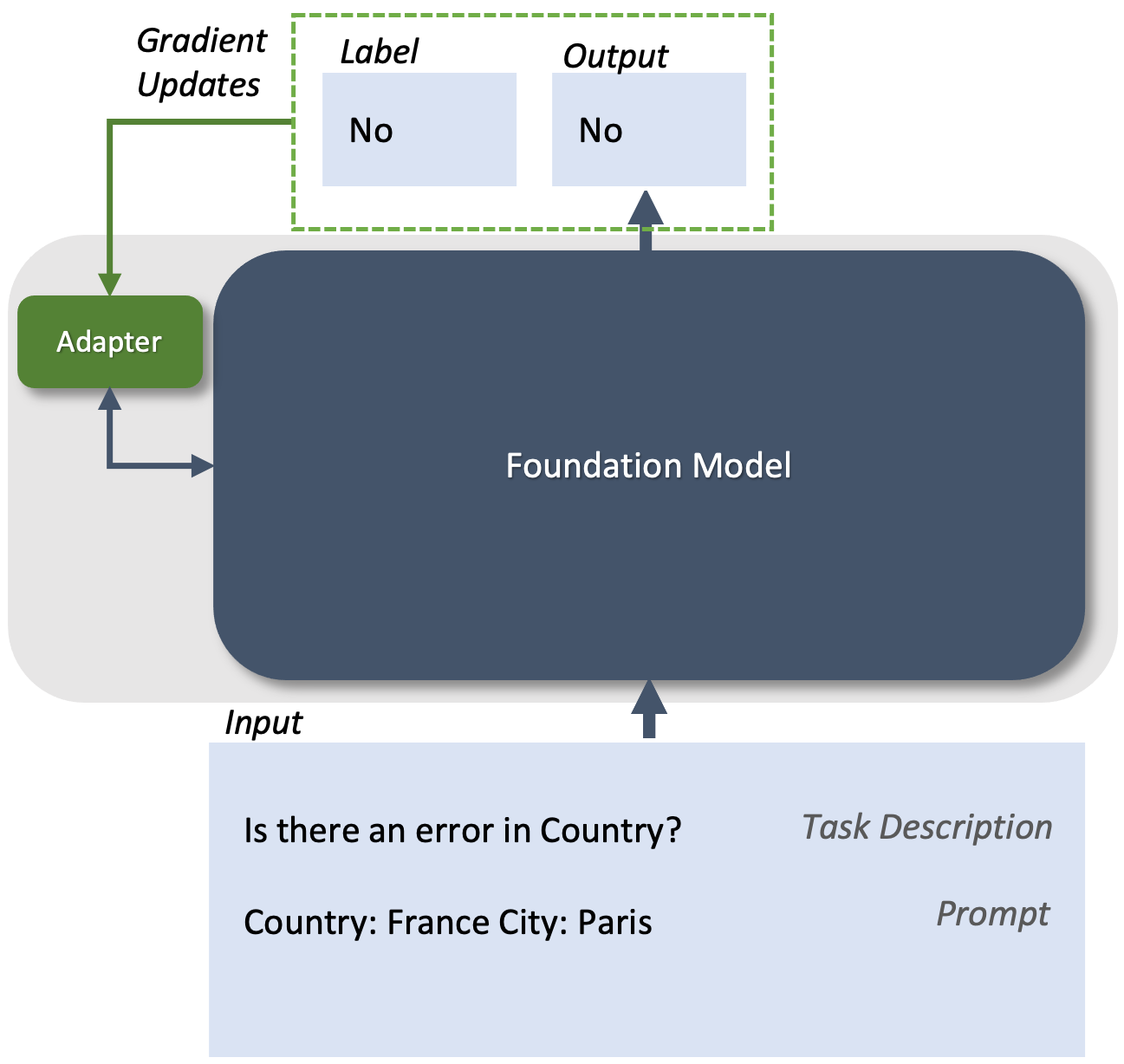}
    \caption{
    Adapter architecture
    }
    \label{fig:adapter}
\end{figure}
\para{Appendix Overview} We structure our appendix into three sections. In ~\cref{appendix:finetune}, we demonstrate how to bridge the performance gap between small \gpt variants (e.g. 1.3B and 6.7B parameters) and the largest \gpt model (175B parameters) through an in-depth discussion of full and lightweight finetuning experiments on data tasks. In ~\cref{appendix:knowledge}, we explore the knowledge encoded  in FMs and the effect of different model update strategies on this knowledge. Finally, in ~\cref{appendix:bias}, we discuss the role of FM bias on data tasks.

\section{Small FM Finetuning Experiments}\label{appendix:finetune}
In this section, we explore ways to bridge the performance gap between the smaller and larger FMs, focusing on (1) full finetuning (updating all parameters) and (2) lightweight finetuning (updating a small number of parameters). 
In~\cref{subsec:finetune_exp}, we evaluate the two approaches and unpack their sample (as measured by number of labeled samples) and training (as measured by number of parameter updates) efficiency tradeoffs.
We find that both lightweight and full finetuning can be used to reduce the performance gap between a 6.7B and a 175B parameter model to an average of 3.1 points.
The lightweight finetuning requires more data than the fully finetuned approach, but updates only 5\% of the model parameters.  

\begin{figure}
    \centering
    \includegraphics[width=0.4\textwidth]{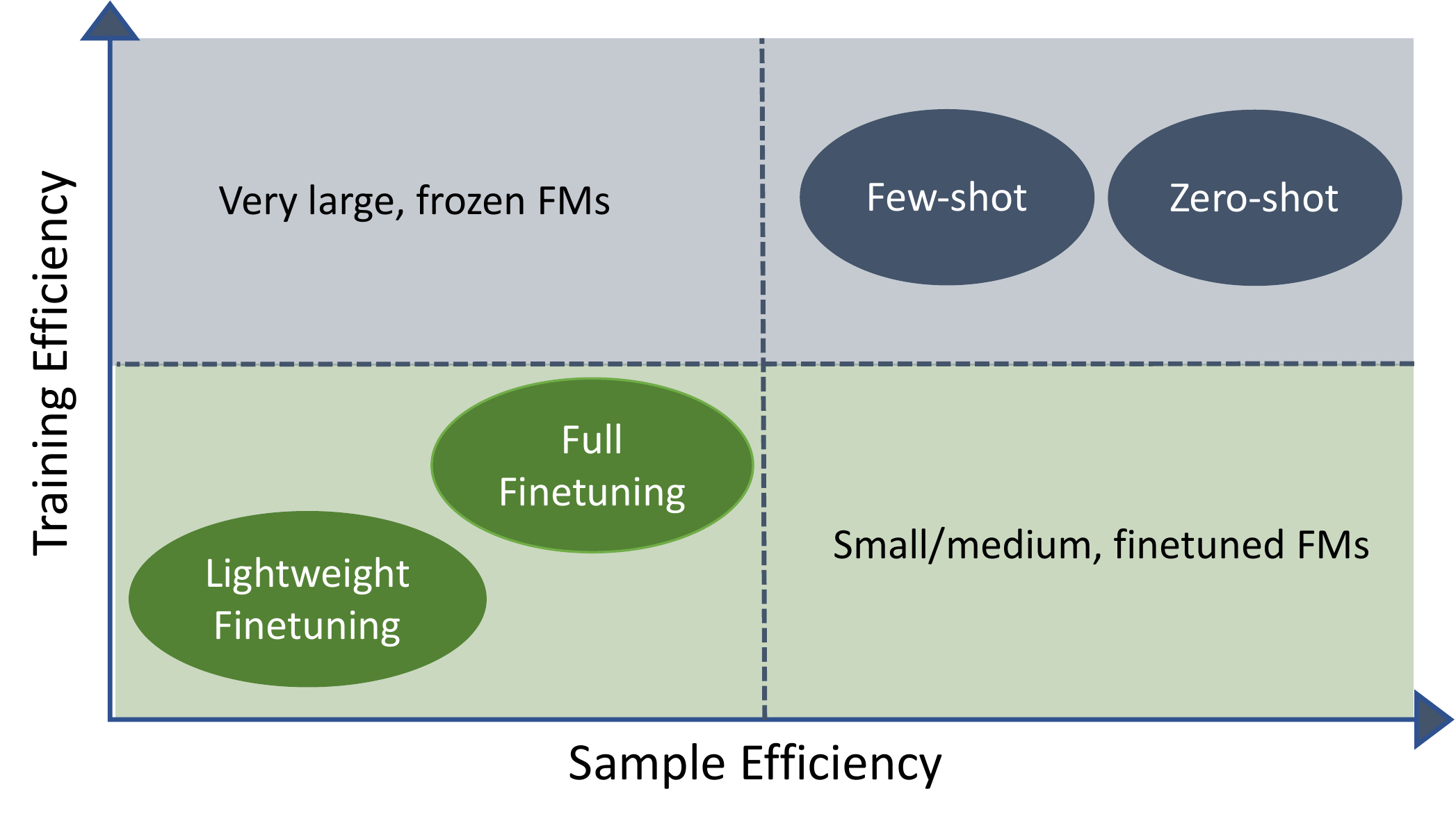}
    \caption{Sample-training efficiency tradeoffs. Larger FMs can be directly used in a zero/few-shot fashion. To achieve similar performance, smaller FMs need additional finetuning.}
    \label{fig:taxonomy}
\end{figure}

\para{\textbf{Full/Lightweight Finetuning}}
The typical practice for achieving optimal task performance in LLMs is to finetune pretrained models with task-specific data. 
This approach is usually training inefficient as all weights in the model are updated.
Interestingly, FMs have been shown to achieve optimal task performance by simply training a lightweight, non-linear layer (i.e. adaptor)~(\cref{fig:adapter}) on the outputs of a frozen model~\citep{frozen2022, lester-etal-2021-power}. 
This is a training efficient method that has proven to be competitive with fully-finetuned approaches~\citep{frozen2022}. 
However, because this approach trains a layer from scratch, it is usually less sample efficient than the fully finetuned approach.
\cref{fig:taxonomy} visualizes the differences in sample and compute efficiency between the two approaches.

For lightweight finetuning, we recursively join a single frozen FM with a small trainable network (the adapter)~\cite{frozen2022}~(\cref{fig:adapter}). 
The prompt input is first passed to the frozen FM whose output embeddings are transformed by the adapter and then passed again as input embeddings to the frozen FM~\cite{frozen2022}.
The generated output of the second pass yields the final answer of the model.  
In this setup, updates are only made to the adapter, i.e. no updates are made to the FM weights alleviating the need for costly training. 

For both the full and lightweight finetuning, the model is fed the natural language prompt in~\cref{subsec:prompting} as input, and trained to generate the task output (e.g., a Yes/No string, or missing value). 


\subsection{Finetuned Performance of Small FMs}
\label{subsec:finetune_exp}
In this section we explore the fully finetuned and lightweight finetuned performance of two small FMs (namely \gpt-6.7B and \gpt-1.3B).
Our goal is to understand whether finetuning can bridge the performance gap with larger models, and analyze the tradeoffs between sample efficiency and training efficiency. 

\subsection{Experimental Setup}
For adapters, we use the GPT-Neo 1.3B~\cite{gpt-neo} and the 6.7B parameter Neo GPT-J~\cite{gpt-j} models available on HuggingFace. 
All finetuning experiments are run on 4-8 A100 GPU machines using Deepspeed~\cite{rasley2020deepspeed} with ZeRO-2 optimizations. 
We use the AdamW optimizer~\citep{loshchilov2017decoupled} with a $10\%$ learning rate warmup followed by a linear decay. 
We train for a maximum or 30 epochs with a learning rate of 1e-4 for adapters and 2e-5 for full finetuning, and save the best model based on validation metrics.

\subsection{Experimental Results} We first review the finetuning results on Walmart-Amazon (EM), Hospital (ED) and Restaurant (DI).

\begin{figure*}[t]
    \centering
    \includegraphics[width=\textwidth]{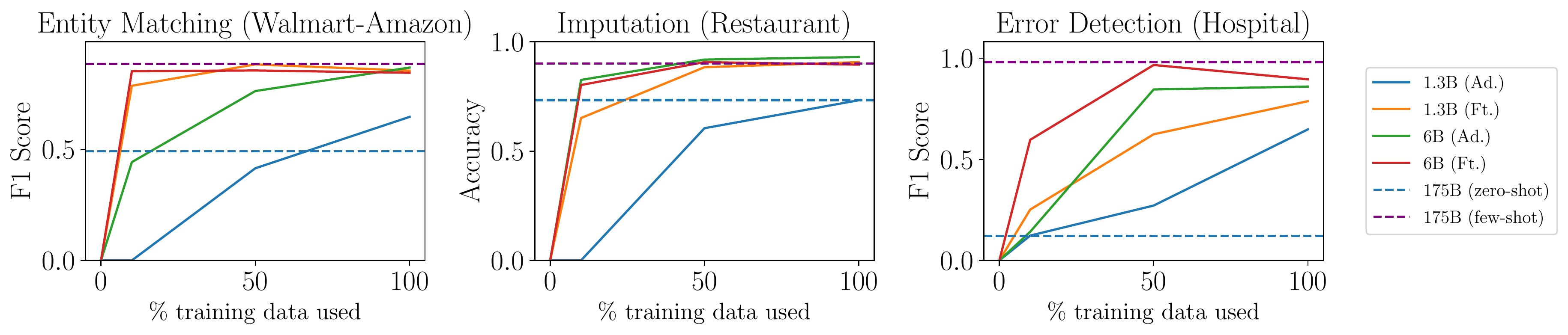}
    \caption{Finetuning experiments: Smaller FMs can be finetuned to bridge the performance gap with larger FMs. Full-finetuing bridges this gap with less data than adapters, but adapters are significantly less expensive to train.}\label{fig:scale-plots}
\end{figure*}

\para{Full Finetuning} Our results show that in the full finetuning setting, we can effectively reduce the performance gap between \gpt-6.7B and \gpt-175B on all datasets~(\cref{fig:scale-plots}), using as little as 10\% of the training set for Walmart-Amazon. 
\gpt-1.3B also matches \gpt-175B performance on Walmart-Amazon and Restaurant, and is within 8  points of the \gpt-175B on Hospital.
Compared to the 6.7B model, the 1.3B model is less sample-efficient and needs more example to bridge the performance gap.

\para{Lightweight Finetuning}
In the lightweight adapter setting, we find that we can effectively bridge the performance gap between \gpt-6.7B and \gpt-175B on both Restaurant and Walmart-Amazon, but not Hospital.
For \gpt-1.3B, we are less effective in bridging the gap, and see an average 25 point performance difference between the \gpt-175B across the three datasets. 
Furthermore, we find that the adapter model outperforms the fully-finetuned approach on two datasets with up to 4 points.

\subsubsection{Discussion} These results show the feasibility of reducing the performance gap between  \gpt-175B and the smaller models. 
We observe some clear sample-training efficiency tradeoffs between the fully and lightly finetuned approaches, which we discuss next. 

\para{Sample Efficiency} 
We find that fully finetuned models are usually more sample efficient than adapters, which could be explained by the need to train a layer from scratch in adapters. 
We also observe that sample efficiency improves with size as the performance of the \gpt-6.7B trained with 10\% of the data is equal to, or better than, the performance of the \gpt-1.3B trained with 50\% of the data in both the full and lightweight finetuning settings. 

Finally, we comment on the performance difference between 6.7B adapter model and the 6.7B finetuned model on Hospital. 
We postulate that adapter model performed poorly on this dataset because the training set was particularly small (100 samples) which suggests that the adapter set-up requires more samples to learn generalizable patterns.  

\para{Training Efficiency} 
We find that the adapter approach for \gpt-6.7B \textit{outperforms} the full finetuning approach on 2 of 3 datasets with 5\% of the amount of trainable parameters, demonstrating that training efficiency need not be sacrificed for quality at this scale. 
However, at the smaller scale (\gpt-1.3B), there is a non-trivial tradeoff between performance and training efficiency suggesting that training efficiency decreases as model size decreases.

\section{Knowledge Ablations}
In this section we seek to explore the encoded FM knowledge that is useful for data tasks. We begin with a qualitative analysis (\cref{appendix:qual_analysis}) and follow with a slice-based analysis that unpacks the impact of different training mechanisms on FM knowledge (\cref{appendix:slice_analysis}).
\label{appendix:knowledge}

\subsection{Qualitative Examples}
\label{appendix:qual_analysis}

To better illustrate the notion of \textit{encoded knowledge} that large FMs possess, we conduct a qualitative analysis where we inspect the model predictions for the task of inferring missing zipcodes or cities given some context~(\cref{tab:qualitative_analysis}) . 
\gpt-175B is able to effectively apply its understanding of functional dependencies between address and zip code and dependencies between address, phone number and city to correctly impute the desired value. 
We further notice that while the smaller models fail to impute the correct values, their generated outputs have the correct semantic type of the missing attribute without out any demonstrative examples.

\begin{table}[t]
    \centering
    \resizebox{0.46\textwidth}{!}{\renewcommand{\arraystretch}{0.95}
    \begin{tabular}{ |p{0.4\linewidth} ||c|c|c| }
     \hline
     \gpt Model (\% data) & freq $=$ 0 & 0 $<$ freq $\leq$ 10 & freq $>$ 10\\
     \hline
     175B (few-shot) & \textbf{100}  & 0.0 & 93.7 \\
     \hline
     6.7B (adaptor, 100\%) & 0.0 &
     \textbf{50.0} & \textbf{98.7} \\\hline
     6.7B (adaptor, 50\%) & 0.0 & 25.0 & 98.7\\
     \hline
     6.7B (adaptor, 10\%) & 0.0  & 0.0 & 87.3\\
     \hline
     6.7B (finetune, 100\%) & 0.0  & 25.0 & 96.2\\
     \hline
     6.7B (finetune, 50\%) & 0.0  & 0.0 & 98.7\\
     \hline
     6.7B (finetune, 10\%) & 0.0  & 0.0 & 89.9\\
     \hline
     
    \end{tabular}}
\caption{Impoverished city entity slice analysis: with lightweight finetuning, smaller model variants perform better than \gpt-175B on infrequently occurring entities.} 
\label{tab:slice_analysis}
\end{table}

\begin{table}[t]
    \centering
    \resizebox{0.46\textwidth}{!}{\renewcommand{\arraystretch}{0.95}
    \begin{tabular}{ |p{0.4\linewidth} ||c|c|c| }
     \hline
     Input Prompt   & GPT3-175B &GPT3-6.7B&GPT3-1.3B\\
     \hline
     ``Address: 1720 university blvd State: AL ZipCode?''  & \textcolor{teal}{$32533$}  & \textcolor{red}{35205}&\textcolor{red}{35901}\\\hline
     ``Address: 26025 pacific coast hwy Phone number: 310/456-5733 City?''  & \textcolor{teal}{Malibu}  & \textcolor{red}{Torrance}&\textcolor{red}{Pacific Beach}\\\hline
     `"Address: 804 north point st Phone number: 415-775-7036 City?"''  & \textcolor{teal}{San Francisco}  & \textcolor{teal}{San Francisco}&\textcolor{teal}{San Francisco}\\

     \hline
    \end{tabular}}
\caption{FMs have an inherent understanding of conditional functional dependencies that are traditionally hard-coded in error detection systems. GPT3-175B is able to accurately map addresses to zip codes and cities.} 
\label{tab:qualitative_analysis}
\end{table}

\subsection{Slice analysis}
\label{appendix:slice_analysis}
We tease apart the differences in encoded knowledge across the different training regimes through a slice-based performance analysis on the Restaurant dataset. 
We use the frequency counts of city names in the training set to define three frequency-based subclasses~(\cref{tab:slice_analysis}). 
We find that for entities that do not occur in the training set, \gpt-175B is the only model that is able to correctly impute the missing values. 
Moreover, we observe that for rare entities---a city value of West LA---these pattern can only be learned through full or lightweight finetuning. 
Interestingly, we find that the lightweight approach more accurately learns the rare subclasses relative to the fully-finetuned approach in both the 100\% and 50\% training set regime. 
We hypothesize that this is because the adapter model is smaller, and is thus less prone to overfitting the training set. \section{Effects of FM Bias}
\label{appendix:bias}
Because FMs are pretrained on large corpuses of web text, they inherit the biases of the data they are trained on. As a result, FMs have been found to contain a number of social biases (e.g. gender, religion, race and more)~\citep{abid2021large, lucy2021gender}. It is important to recognize the effects of these biases when utilizing FMs in downstream applications. When applying FMs to data management tasks, we need to be wary of the fact that the inherent biases of these models may be propagated through their actions. Concretely, for data repair, the encoded biases of the FM may cause it to replace or correct entries in a biased manner (e.g. adding a suffix of Ms. to a traditionally female name). As a result, we need mechanisms for mitigating model bias and monitoring for bias in model outputs in deployment settings. Recent work takes a first step in this direction by proposing methods for automatically detecting bias-sensitive tokens and correcting these tokens to mitigate biases~\cite{liang2021towards}.

\section{Evaluating More FMs on Data Wrangling Tasks}\label{appendix:benchmarking} This works focuses on evaluating \gpt---one of the many large FMs---on a collection of data integration and cleaning tasks. We contribute tasks from this paper to the HELM benchmark~\citep{liang2022holistic}, which evaluates performance of a broader set of large FMs on these tasks. 

\end{document}